# Morphological annotation of Korean with Directly Maintainable Resources

**Ivan Berlocher[1], Hyun-gue Huh[2], Eric Laporte[2], Jee-sun Nam[3]**

[1] Saltlux
#5-6F, Deok-il Bldg., 967, Daechi-dong, Gangnam-gu, Seoul
135-848, Korea
{ivan@mobico.com}
[2] Université de Marne-la-Vallée, IGM
5, bd Descartes, 77454 Marne-la-Vallée CEDEX 2, France
{hyungue.huh@univ-mlv.fr} {eric.laporte@univ-mlv.fr}
[3] Hankuk University of Foreign Studies, DICORA
89 Wangsan-ri Mohyun-myun
Yongin-shi Kyunggi-do
449-791, Korea
{namjs@hufs.ac.kr}

**Abstract**
This article describes an exclusively resource-based method of morphological annotation of written Korean text. Korean is an agglutinative language. Our annotator is designed to process text before the operation of a syntactic parser. In its present state, it annotates one-stem words only. The output is a graph of morphemes annotated with accurate linguistic information. The granularity of the tagset is 3 to 5 times higher than usual tagsets. A comparison with a reference annotated corpus showed that it achieves 89% recall without any corpus training. The language resources used by the system are lexicons of stems, transducers of suffixes and transducers of generation of allomorphs. All can be easily updated, which allows users to control the evolution of the performances of the system. It has been claimed that morphological annotation of Korean text could only be performed by a morphological analysis module accessing a lexicon of morphemes. We show that it can also be performed directly with a lexicon of words and without applying morphological rules at annotation time, which speeds up annotation to 1,210 word/s. The lexicon of words is obtained from the maintainable language resources through a fully automated compilation process.

## 1. Introduction

In the first phase of the processing of a written text, words are annotated with basic information, such as part of speech (POS). Our objective was to explore methods likely to enhance the performances of state-of-the-art systems of Korean morphological annotation as regards two criteria of quality: the accuracy and informative content of output; the ability of underlying language resources to undergo an evolution in a controlled way.

Korean is an agglutinative language. A Korean word (*eojeol*) usually consists of an undelimited concatenation of several morphemes: one or several stems, followed by zero, one or several functional morphemes. For example, the word 컸다:*kôssta* 'was big' has a stem, ㅋ:*k-* 'big', and two functional morphemes, -었:-*ôss-* (past) and -다: -*ta* (declarative). We will call stems all lexical morphemes, as opposed to functional morphemes. The surface form of a morpheme occurring in a text may depend on neighbouring morphemes and differ from its base form or lexical form. For example, the surface form of the stem 'big' above is ㅋ:*k-* before the suffix -었:-*ôss-* (past), but its base form is 크:*keu-*. The variations are termed as phonotactic, and the surface variants are called allomorphs. The objective of morphological annotation is to identify morphemes and assign relevant information to them.

In Section 2, we outline the state-of-the-art approach to Korean morphological annotation. Section 3 explains the objectives of our work. In Section 4, we describe our methods of management of the underlying language resources and the operation of the annotator. Section 5 reports our evaluation experiments. A conclusion and perspectives are presented in Section 6.

## 2. State of the art

Several morphological annotators of Korean text are available. The Hangul Analysis Module (HAM[1]) is one of the best Korean morphological analysers. Other fairly representative examples are described in (Park et al., 1998) and in (Cha et al., 1998; Lee et al., 2002). In most morphological annotators of Korean, the output for each morpheme is presented in two parts: the morpheme itself, and a grammatical tag. Morphemes are usually presented in their base form if they are stems, and in they surface form if they are functional morphemes. The operation of these annotators generally involves frequency-based learning from a tagged corpus with statistical models (Park et al., 1998; Lee et al., 2002). They typically report that 95% to 97% of morpheme-tag pairs are correct.

The current state of the art is largely sufficient for natural language processing (NLP) applications such as web searching, text retrieval and text clustering. However, more ambitious NLP applications such as translation, text generation and text-to-speech synthesis will require more accurate annotation. Firstly, they will require higher recall, if recall is defined in the same way as answer inclusion rate (AIR) in (Lee & Rim, 2005): a 97% recall means a 10-word sentence has a 30% probability that one of its words has not been assigned its correct tag, which will make syntactic parsing practically impossible. Secondly, ambitious applications will require more informative tags: the tagsets currently in use do not contain the syntactic and semantic features required to resolve ambiguity and recognise sentence structures. Examples of such features are the number of essential complements of verbs and the postpositions expected in these complements. Thirdly, most annotators provide the base form of stems, but not of functional morphemes.

---
[1] http://nlp.kookmin.ac.kr/HAM/kor/ham-intr.html

Upgrading the state-of-the-art corpus-driven systems in order to achieve more accurate annotation would involve shifting to tagsets of a higher granularity, and consequently constructing new annotated corpora. Due to data sparseness, training would presumably imply much larger annotated corpora in order to achieve the same recall values, and *a fortiori* to enhance them.

## 3. Objectives

The alternative approach based on directly maintainable language resources is complementary to the corpus-driven approach, because it behaves gracefully with fine-grained tagsets and does not require annotated corpora. This approach might eventually lead to hybrid systems able to outperform present annotators. The resource-based approach to annotation can be defined by two features:
- the quality of results depends essentially on the quality of manually constructed language resources;
- the maintenance of the annotator is performed through the direct, linguistically motivated maintenance of these resources.

We undertook this work in order to investigate a purely resource-based approach to morphological annotation of Korean. We developed a system with a more informative tagset than current systems: 150 tags vs. 30 to 60. Our objectives were to assess:
- the recall achieved by such a system;
- the level of readability and of flexibility of the language resources, a significant criterion of quality which is often overlooked in literature.

In addition, our systems operates by searching directly a lexicon of words, instead of running a morphological analysis module accessing a lexicon of morphemes. Our solution, which was previously deemed unfeasible (Han & Palmer, 2005), speeds up annotation, since all computation related to morpheme agglutination is performed in advance.

## 4. Methodology

The number of Korean words is in the hundreds of millions. In order to construct and handle an actual lexicon of words, we resorted to classic techniques of lexicon compression and lookup (Appel & Jacobson, 1988; Lucchesi & Kowaltowski, 1993; Revuz, 1992; Silberztein, 1991), but the application to an agglutinative language involved implementing new software components, which received an open-source status[2] and were integrated to the open-source Unitex system [3] (Paumier, 2002).

Our tag set comprises 150 tags. Comparable figures are 47 (Chae & Choi, 2000), 33 (Lee et al., 2002) and 29 (Han & Palmer, 2005). Our tag set is therefore more fine-grained and tags are more informative. In addition, the tags are structured. They combine a general tag taken in a list of 16 general tags, and 0 to 4 features specifying subcategories. There are 91 features with a total of 152 values. This structure is in conformity with emerging international standards in representation of lexical tags (Lee et al., 2004).

The final part of some verbal and adjectival stems undergoes phonotactic variations when a suffix is appended to them. For example, the stem *keu*- 'big' becomes *k*- before the suffix -*ôss*- (past). Stem allomorphs are generated from base-form stems with 71 finite-state transducers of the same type as those used to inflect words in inflectional languages (Silberztein, 2000). A finite-state transducer is a finite-state automaton with input/output labels. The input part of the transducers specifies letters to remove or to add in order to obtain the allomorph from the base form. The output part specifies the tag and compatibility symbol (see below) to be assigned to the allomorph. These transducers are viewed and edited in graphical form with Unitex.

The combination of a stem with a sequence of suffixes obeys constraints. Checking these constraints is necessary to discard wrong segmentations. We distinguish two types of suffixes: derivational and inflectional.

Derivational suffixes are markers of verbalization, adjectivalization and adverbialization. They are appended by applying transducers of the same type as above. In our current version, 8 transducers append derivational suffixes. These transducers invoke 5 subgraphs, thus constituting recursive transition networks (RTN).

Inflectional suffixes comprise all other types of suffixes. A single (possibly derived) stem can be combined with up to 5,500 different sequences of inflectional suffixes. Compatibility between stems and inflectional suffixes is represented by a set of 59 compatibility symbols (CS). Each stem and stem allomorph is assigned a CS, which defines the set of suffix sequences that can be appended to it. The CSs have been semi-automatically inserted by Huh Hyun-gue into the lexicons of stems. For this work, we did not make any attempt at recognising multi-stem words, i.e. words with an undelimited sequence of stems.

The standard model for representing phonotactic and grammatical constraints is the finite-state model. In our system, constraints are represented in finite-state transducers. These transducers describe sequences of suffixes. Their input represents surface forms and their output represents base forms and tags. For readability, they are edited and viewed graphically, and since most of the transducers are large and would not display conveniently on a single screen or page, they take the form of RTNs: transitions can be labelled by a call to a sub-transducer. Most of the sub-transducers that they call are shared, which reduces the level of redundancy of the system. The total number of simple graphs making up the RTNs is 230.

The directly maintainable language resources of our system are thus the lexicon of stems (Nam, 2002, 2003, 2004), the allomorph transducers and the suffix RTNs. Samples of all these resources are distributed under the LGPL-LR license[4]. The lexicon of words is compiled from these resources into a data set with an index for fast matching. This index is a finite-state transducer over the Korean alphabet of letters. The compilation of the lexicon of words follows several sequential steps. First, all resources are converted from the Korean syllabic alphabet to the Korean alphabet of letters. In a second step, we generate lexicons of stem allomorphs and of derived stems from the base-form stem lexicons by applying the allomorph transducers with Unitex. In a third step, each resulting lexicon of stems is compiled by the Unitex lexicon compiler into a finite-state transducer. The final states of the stem transducers give access to the lexical information, and in particular to the CSs of the stems. In a fourth step, each RTN of sequences of suffixes is converted into a list by a path enumerator, and each of these lists is processed by the lexicon compiler into a finite-state transducer. The names of the ending transducers contain the corresponding CSs. In the final step, the stem transducers and the ending transducers are merged into a word lexicon, which is also represented as a finite-state transducer. This operation links the final states of the stem lexicons to the initial states of the corresponding ending lexicons. The path enumerator and the lexicon link editor have been implemented for this annotator. The path enumerator allows for breaking cycles in the graph of calls to sub-transducers, so that the enumeration remains finite.

---

[2] These software components are distributed under the Lesser General Public License (LGPL): http://www.gnu.org/copyleft/lesser.html.

[3] http://www-igm.univ-mlv.fr/~unitex/manuelunitex.pdf

[4] The Lesser General Public License for Language Resources, http://www-igm.univ-mlv.fr/~unitex/lgpllr.html, is approved by the Free Software Foundation, http://www.fsf.org.

These operations are independent of the text to be annotated; they are performed beforehand. They need to be repeated when one of the resources is updated.

The operation of the morphological annotator is simple. The text is tokenised (words are tokens) and the lexicon of words is searched for the words. Our system presents its output in an acyclic automaton[5] of morphemes, as in Lee et al. (1997), but displayed graphically.

The output for each morpheme is presented in three parts: surface form, base form, and a structured tag providing the general tag and syntactic-semantic features. Word separators such as spaces are also present in this automaton.

## 5. Evaluation

We evaluated our system by running it on a subset of the annotated corpus of Korean constructed at KAIST in the framework of the KIBS project (Chae & Choi, 2000). This corpus has been tagged at KAIST by a stochastic tagger, and then manually corrected during 3 years. We used a 618,232-word sub-corpus. The raw text occupies 8.7 Mb in UTF16-LE.

### 5.1. Efficiency

The complete annotation process: sentence segmentation, tokenization, alphabet conversion, lexicon search, alphabet conversion, sorting and saving, took 509 s on a Pentium M 1.2 GHz with 1Gb memory, i.e. 1210 word/s.

### 5.2. Recall

We assessed the quality of the output of our morphological annotator by comparison with the KAIST annotated sub-corpus. This corpus has not been used during the construction of the resources or of the software.

---
[5] Also called a directed acyclic graph (DAG) or, improperly, a lattice.

Our tag set is much more fine-grained than that of the tagged corpus: 150 word tags vs. 47. We had to downgrade the output of Unitex to be able to compare it with the reference corpus. Our tags were mapped onto the KAIST tags by abstracting away the information not provided in the reference tagset. Thus, the comparison does not take into account all the information in the output of our system[6].

As we did not use any language resources for Korean compound words in our annotator, we restricted the comparison to the 447,835 words annotated with a single stem in the reference corpus. The tag in the reference corpus was present in the output of Unitex in 89% of these words, which means the recall or AIR is 89%.

In fact, this value is not an accurate measure of the recall of our system. We examined a few words for which the reference tag is absent from the output of Unitex. They can be distributed in 3 cases:

(i) Cases where the reference tag is correct, but Unitex does not assign it because of an error in our language resources. Example: the token 라는:*laneun* 'called' is annotated {라,라.ns_inv} {는,는.npost+au_the} 'the A note' whereas it consists of a single suffix, correctly tagged {라는.exm} in the reference corpus.

(ii) Cases where the reference tag can be considered correct, but one of the tags in the Unitex output is also correct, though different from the reference tag, due to different guidelines. Examples:그런:*geulôn* 'such' is seen as a demonstrative determiner and tagged {그런.md} in the reference corpus, but seen as an inflected adjective and tagged {그러,그렇다.re} {ㄴ,는.apost+de_pre} by Unitex. In particular, the Unitex morpheme segmentation is generally finer, e.g. in 짐승만도:*jimseungmando* 'even like a beast', tagged {짐승.nc} {만도.jx} in the reference corpus, but {짐승,짐승.co_cnt} {만,만 .npost+au_lim} {도,도.npost+au_add} by Unitex.

(iii) Cases where one of the Unitex tags is linguistically more motivated than the reference tag. Example:표리:*pyoli* 'be a figure', tagged {표리.nc} {를.jc} 'appearances and reality' in the reference corpus, would have been better described by {표.nc} {리.ef} which is the equivalent of one of the Unitex analyses.

In other words, the choice of a reference corpus was necessary for our evaluation experiment, but the guidelines and descriptive choices in the reference corpus may have an equal, higher or lower linguistic motivation as compared to ours.

This 89% recall value is encouraging. Firstly, the errors and lexical lacunae in the language resources can be corrected over time. Secondly, this result has been obtained without any corpus-driven learning: by combining it with the output of a state-of-the-art tagger where Unitex does not find any analysis, a much higher recall would be easy to obtain.

### 5.3. Precision

The evaluation of the precision of the output of our annotator was not one of our main objectives, since ambitious NLP applications such as translation, text generation and text-to-speech synthesis require the operation of a syntactic parser, which can resolve residual lexical ambiguity. The recognition of phrases and other linguistic patterns with several words does not require high precision either.

---
[6] Of course, we knew it would be so beforehand, but we created this fine-grained tagset anyway because no such tagset was available and our objective of enhancement of methods is of higher significance than the straightforwardness of comparison between systems.

However, for the sake of completeness, we measured precision as the average value of the proportion of correct tags among the tags retained by Unitex for a word. The number of correct tags for a word is evaluated by comparison with the reference corpus, which yields for each word either 0 or 1. The value obtained is 32%. Ambiguity resolution techniques can be applied to the output of our annotator: the syntactic approach would take advantage of the rich linguistic information provided in output; classic statistical approaches and priority rules (Kang, 1999) are applicable as well.

### 5.4. Flexibility

Flexibility of language resources of NLP systems is often overlooked in literature, but it is a significant criterion of quality, because the extension of a system to input texts of new types or of a new period of time is a frequent task.

For annotators, this criterion of quality is not numerically assessable. We invite the reader to have a look at the free sample of directly maintainable language resources distributed with Unitex and to check that they are available in readable, editable forms: lists of stems, graphs of suffixes. The possibility of explicitly updating these resources allows users to extend them, to correct errors in the output, and to control the evolution of the system, which has thus a satisfactory level of flexibility.

In contrast, a corpus-based annotator is adapted by feeding new corpora into the training process. Extending it involves the costly task of tagging a corpus of new texts. In addition, the correction of an error in its corpus-trained behaviour involves a new training, but nothing ensures that the new training will correct the error and not bring about new errors. Such systems have therefore a limited flexibility.

## 6. Final Remarks

This work opens several perspectives. The resources will be extended by running the annotator and analysing output. Additional experiments will be carried out to investigate the annotation of compound words, using both lexicons of compounds (Bae, 2001) and rules of combination of stems. Existing approaches to the analysis of unrecognised morphemes could be combined to our system: they are complementary to our resource-based approach, and would take advantage of the rich information provided on the neighbouring words. Finally, parallel systems are under construction for Finnish and Hungarian, two other agglutinative languages with undelimited morphemes.

## Acknowledgments

This work was partially supported by the Outilex project and by the CNRS.